\documentclass[journal]{IEEEtai}
\usepackage{amsmath,amsfonts}
\usepackage{algorithmic}
\usepackage{algorithm}
\usepackage{array}
\usepackage[caption=false,font=normalsize,labelfont=sf,textfont=sf]{subfig}
\usepackage[colorlinks,urlcolor=blue,linkcolor=blue,citecolor=blue]{hyperref}
\usepackage{textcomp}
\usepackage{stfloats}
\usepackage{url}
\usepackage{verbatim}
\usepackage{cite}
\usepackage{multirow}
\usepackage{multicol}
\usepackage{booktabs}
\usepackage{siunitx}
\usepackage{listings}
\usepackage{color,array}
\usepackage{graphicx}

\newtheorem{theorem}{Theorem}

\begin{document}

\title{FAST: Feature Aware Similarity Thresholding for weak unlearning in black-box generative models}

\author{Subhodip Panda, Prathosh AP \\
Indian Institute of Science}


\maketitle

\begin{abstract}
 
The heightened emphasis on the regulation of deep generative models, propelled by escalating concerns pertaining to privacy and compliance with regulatory frameworks, underscores the imperative need for precise control mechanisms over these models. This urgency is particularly underscored by instances in which generative models generate outputs that encompass objectionable, offensive, or potentially injurious content. In response, \emph{machine unlearning} has emerged to selectively forget specific knowledge or remove the influence of undesirable data subsets from pre-trained models. However, modern \emph{machine unlearning} approaches typically assume access to model parameters and architectural details during unlearning, which is not always feasible. In multitude of downstream tasks, these models function as black-box systems, with inaccessible pre-trained parameters, architectures, and training data. In such scenarios, the possibility of filtering undesired outputs becomes a practical alternative. Our proposed method \emph{Feature Aware Similarity Thresholding (FAST)} effectively suppresses undesired outputs by systematically encoding the representation of unwanted features in the latent space. We employ user-marked positive and negative samples to guide this process, leveraging the latent space's inherent capacity to capture these undesired representations. During inference, we use this identified representation in the latent space to compute projection similarity metrics with newly sampled latent vectors. Subsequently, we meticulously apply a threshold to exclude undesirable samples from the output. Our implementation is available at \hyperlink{here}{https://github.com/Subhodip123/weak-unlearning-gan}
\end{abstract}

\begin{IEEEImpStatement}
The primary goal of this study is twofold: first, to elucidate the relationship between filtering and unlearning processes, and second, to formulate a methodology aimed at mitigating the display of undesirable outputs generated from black-box Generative Adversarial Networks (GANs). Theoretical analysis in this study demonstrates that, in the context of black-box models, filtering can be seen as a form of weak unlearning.  Importantly, our proposed methodology exhibits efficacy even in scenarios where the user feedback exclusively consists of negative samples, devoid of any positive examples. The veracity and utility of our approach are empirically substantiated through a series of experiments conducted within a low data ($<100$ samples) regime, featuring the utilization of DC-GAN and Style-GAN2 architectures on the MNIST and CelebA-HQ datasets respectively. 
\end{IEEEImpStatement}


\section{Introduction}
Deep generative models have gained widespread popularity for the generation of authentically realistic synthetic data. These models find utility in numerous domains where access to authentic data is limited, exemplified by their application in fields such as medical imaging \cite{med-survey1,med-survey2}, remote sensing \cite{rs-survey1,rs-survey2}, hyper-spectral imagery \cite{hs-survey1,hs-survey2} and others\cite{gen-survey1,gen-survey2,gen-survey3} and also in various downstream tasks, including but not limited to classification \cite{gen-classification1,gen-classification2}, segmentation \cite{gen-seg} etc. However, despite their versatile applicability, the presence of undesirable samples within the dataset and the inherent biases ingrained in these datasets \cite{dataset-bais} contribute to instances where these models produce outputs that are regarded as objectionable, offensive, or even potentially harmful. Thus, recent regulatory measures related to data privacy and protection, such as the European Union's General Data Protection Regulation (GDPR) \cite{gdpr} and the California Consumer Privacy Act (CCPA) \cite{ccpa}, compel organizations to enforce stringent controls on models that have the potential to generate content deemed harmful or offensive. To address this challenge, our focus lies in exploring methodologies that prevent the display of specific output types characterized by undesirable features. In a parallel vein of inquiry, recent research in the domain of \emph{machine unlearning} within generative models \cite{gan-unlearn, diffusion-unleanrn} has delved into the concept of forgetting, wherein the primary objective revolves around the modification of generative model parameters to ensure the non-generation of undesirable images. It is worth noting that all of the aforementioned studies operate under the assumption that, during the unlearning phase, accessibility to the learned model's architecture, parameters, and, at times, even the training dataset is readily available. However, there exist scenarios wherein these models are employed as service providers for diverse downstream tasks, functioning as black-box models. In this more stringent context, the pertinent model parameters, architectures, and the underlying dataset remain entirely obscured from the user's purview. In such a constrained environment, conventional unlearning procedures or the prospect of retraining the model from its inception becomes an unattainable endeavor due to the complete lack of access to model parameters and architectures. Consequently, the question that necessitates elucidation is as follows: \\

\begin{centering}
    \textit{What are the feasible remedies that can effectively suppress the display of undesired outputs when the model is entirely opaque, operating as a black-box system?}    
\end{centering} \\

In a black-box scenario, filtering where the task involves screening out undesired outputs while retaining other outputs, is a viable practical solution. This research addresses the formidable challenge of effectively filtering undesired outputs generated by black-box generative models within the feedback-based unlearning framework. Specifically, the user is presented with a curated collection of generated samples stemming from a pre-trained black-box Generative Adversarial Network (GAN). Within this assortment, the user is tasked with distinguishing a subset as undesirable (negative) and another subset as desirable (positive). This task is subject to constraints, notably a limited sample size for the user feedback (typically not exceeding 100 samples). While the concept of employing a binary classifier for filtering appears straightforward, the crux of the challenge resides in training a classifier for this task. The primary hindrance is the scarcity of available user feedback samples, making the task inherently intricate. Moreover, conventional filtering mechanisms, which utilize such classifiers, lack awareness of the underlying representation of the undesired feature. Consequently, the development of filtering mechanisms equipped with the capacity to grasp the representation of undesired features becomes a complex and non-trivial endeavor. Adding to the complexity is the requirement to construct a comprehensive and informative representation of the undesired feature. This representation must be meticulously designed to exclusively encapsulate the distinctive characteristics of the undesired feature, without incorporating any extraneous information related to correlated features. The entanglement between the representations of semantic features further complicates this task. For instance, within the CelebA-HQ~\cite{celebA} dataset, the beard feature exhibits substantial correlations with attributes like mustache and male gender. Therefore, an effective filtering mechanism must be equipped with the ability to discern and harness a representation that singularly encapsulates the beard feature, while prudently avoiding the inclusion of information about other correlated attributes such as mustache and gender.

In this work, we introduce an innovative family of filtering mechanisms, collectively referred to as \emph{Feature Aware Similarity Thresholding (FAST)} meticulously designed to proficiently detect and eliminate outputs featuring undesired attributes while inherently possessing an awareness of the representation of these undesired features. Our methodology, reliant upon user-provided positive and negative samples, effectively exploits the latent space to characterize the representation of the undesired feature. The fundamental rationale underpinning our approach centers on the latent space's remarkable capacity to encode intricate and interpretable representations of semantic features. This encoded representation encapsulates the essential traits of the undesired feature, thereby endowing the system with the ability to effectively distinguish it from other attributes. The operational framework of our method entails an assessment of the similarity between newly sampled latent vectors and the previously identified latent feature representation corresponding to the undesired attribute. Leveraging the principle that latent vectors encoding similar features are proximate in the latent space, we gauge the alignment between the latent vectors and the undesired feature, facilitating the precise identification of undesired outputs. Subsequently, a well-considered threshold is applied to this similarity metric, leading to the segregation and exclusion of outputs that manifest the undesired feature.

We summarize our contribution as follows:
\begin{itemize}
    \item This work represents the first instance where filtering is introduced as a form of weak unlearning to block undesired features within black-box Generative Adversarial Networks (GANs).
    \item  Theoretical analysis conducted in this research establishes that, in the context of black-box generative models, the processes of filtering and weak unlearning are equivalent.
    \item Our proposed filtering mechanism relies on user-provided feedback, encompassing undesired (negative) and desired (positive) outputs. Acknowledging the potentially demanding nature of user feedback, our method is tailored to operate effectively within a few-shot setting, wherein the number of user feedback samples does not exceed 100. Furthermore, our approach demonstrates its robustness by functioning efficiently even when the user provides only negative samples, without including any positive samples. In addition to its efficacy in filtering undesired outputs, our method distinguishes itself by its intrinsic awareness of the representation of undesired features.
    \item We assess and validate the performance of our proposed method across various GAN architectures, including DC-GAN and Style-GAN2, employing datasets such as MNIST and CelebA-HQ, thereby showcasing its effectiveness and versatility in different scenarios and datasets.
\end{itemize}

\section{Background}

\subsection{Machine Unlearning}
\emph{Machine unlearning}~\cite{unlearningsurvey1,unlearningsurvey2} refers to the process of deliberately forgetting specific acquired knowledge or erasing the impact of particular subsets of training data from a trained model. Naive unlearning methods typically entail the removal of undesirable data from the training dataset, followed by retraining the model from scratch. However, this approach becomes computationally prohibitive when unlearning requests are made iterative for individual data points. Inspired by concerns surrounding privacy protection,~\cite{forget4} introduced methods for data deletion within statistical query algorithms, coining the term \emph{machine unlearning}. Unfortunately, these methods are primarily suitable for structured problems and do not extend to complex machine learning algorithms, such as k-means clustering \cite{unlearning4} or random forests \cite{unlearning6}. Efficient deletion algorithms were devised for the k-means clustering problem, which introduced effective data deletion criteria applicable to randomized algorithms based on statistical indistinguishability. Building upon this criterion, \emph{machine unlearning} methods are broadly categorized into two main types: exact unlearning \cite{unlearning4,unlearning6} and approximate unlearning \cite{unlearning7}. Exact unlearning endeavors to completely eradicate the influence of unwanted data from the trained model, necessitating precise parameter distribution matching between the unlearned and retrained models. In contrast, approximate unlearning methods only partially mitigate data influence, resulting in parameter distributions that closely resemble the retrained model, albeit with minor multiplicative and additive adjustments. To eliminate the influence of unwanted data, \cite{influenceremoval} proposed technique employs parameter perturbation based on cached gradients, offering computational efficiency while increasing memory usage. Other methods \cite{influenceremoval2,influenceremoval3} have suggested the use of influence functions for this purpose. However, these approaches are computationally demanding due to the necessity of hessian inversion techniques and are limited to small convex models. To extend the applicability of influence removal techniques to non-convex models like deep neural networks, a scrubbing mechanism \cite{unlearning8} was introduced within a classification framework. However, until recently, it remained unclear how these techniques could be applied to unsupervised models, particularly state-of-the-art generative models. In response, \cite{unleanring11} introduced an \emph{Adapt-then-Unlearn} mechanism designed to unlearn a GAN in a zero-shot setting where the training dataset is inaccessible. Nevertheless, this method does not readily extend to black-box generative models, presenting a challenge in the context of \emph{machine unlearning} for black-box models.

\subsection{Generative Adversarial Network}
Generative Adversarial Networks (GANs), a prominent class of generative models primarily used for image synthesis, were first introduced in the seminal work by \cite{gan-goodfellow}. The GAN framework consists of two neural networks: a generator network and a discriminator network. GANs are trained through an adversarial loss function, which strives to minimize the divergence between the distribution of original data and that of generated (fake) data. The generator network, once trained, takes random latent vectors sampled from a normal distribution as input and produces synthetic data that closely resembles the original data. This process was theoretically elucidated by \cite{gan-goodfellow,wasserstein-gan}, who demonstrated that the generator aims to minimize the Jensen-Shannon divergence between the distribution of fake data and that of the original data. Moreover, \cite{f-gan} expanded upon this idea by showcasing that various divergence metrics, with the Jensen-Shannon divergence being one of them, can be minimized to facilitate the training of effective generative models. They introduced a broader class of models called f-GANs, with the specific choice of the f-divergence metric influencing the optimization process. To further enhance the capabilities of GANs, \cite{info-gan} proposed an information-theoretic extension known as Info-GAN. This innovation allowed GANs to learn a disentangled latent space, significantly improving the interpretability and editability of the generated data. Addressing challenges related to GAN training stability, \cite{wasserstein-gan} introduced the idea of training GANs with Wasserstein distance metrics, which contributed to more stable and reliable training processes.
Furthermore, the field of generative models has witnessed the introduction of numerous specialized GAN variants\cite{prog-gan} aimed at synthesizing high-quality images. For instance, \cite{style-gan1,style-gan2,style-gan3} proposed a style incorporation technique during model training, achieving state-of-the-art results for generating high-resolution images.

\subsection{Latent Space Representations and Inversion}
The latent representation of Generative Adversarial Networks (GANs) often proves to be valuable for auxiliary supervised downstream tasks. Furthermore, this latent space exhibits the capacity to encode a diverse array of semantic features that find utility in tasks such as image manipulation and editing, among others, as demonstrated by \cite{info-gan}. Numerous studies \cite{style-gan1,style-gan2,style-gan3} have underscored the latent space's remarkable representation capabilities, which encompass a wide range of interpretable semantics. For instance, in the context of Style-GAN2, it was observed that the learned $\mathcal{W}$ space possesses greater disentanglement compared to the random noise space $\mathcal{Z}$. This characteristic enhancement in disentanglement contributes to improved editability and perceptual quality in generated images. Additionally, latent spaces often exhibit commendable clustering properties \cite{cluster-gan}, where data with similar semantic features cluster closely within the latent space. To harness the rich structural and representational attributes of the latent space, it is essential to project data back into this space through a process known as inversion \cite{gan-inversion-survey}. Inverter networks are employed for this purpose, taking data as input and producing corresponding latent vectors, which enable the GAN model to accurately reconstruct the data. Two primary approaches are typically used for inversion: iterative optimization and inference with an encoder \cite{enc-inverter1,enc-inverter2}. Recent developments in encoder-based methods \cite{e4e-inversion, hyper-inverter} have yielded impressive outcomes, not only in terms of reconstruction but also in enhancing expressiveness. 

\section{Theory and Analysis}

The notations are as follows:
\begin{center}
    \begin{itemize}
        \item $\Theta$ := the space of all parameters of generative models
        \item $\mathcal{X}$ := the space of all data points
        \item $\mathcal{Z}$ := space of all latent vectors of generative models
    \end{itemize}
\end{center} 

Consider a specific training dataset $D$ consisting of $m$ i.i.d samples $\{{x_i}\}_{i=1}^m$, drawn from a distribution $P_\mathcal{X}$ over the data space $\mathcal{X}$. A generative model ($G$) trained on $D$ aims to learn the data distribution $P_\mathcal{X}$. The pre-trained generative 
model is denoted as $G_{\theta^{init}}$,  with parameters $\theta^{init} \in \Theta$. Based on the outputs of the generative model, the user designates a portion of the data space as undesired, referred to as $\mathcal{X}_f$. Therefore, the entire data space can be expressed as the union of $\mathcal{X}_r$ and $\mathcal{X}_f$, where $\mathcal{X} = \mathcal{X}_r \bigcup \mathcal{X}_f$. We denote the distributions over $\mathcal{X}_f$ and $\mathcal{X}_r$ as $P_{\mathcal{X}_f}$ and $P_{\mathcal{X}_r}$, respectively. The objective is to create a mechanism in which the generative model does not produce outputs within the domain $\mathcal{X}_f$. This implies that the generative model should be trained to generate data samples conforming to the distribution $P_{\mathcal{X}_r}$. A naive approach to achieving this is by retraining the entire model from scratch using a dataset $D_r$. Here, $D_r$ is defined as the intersection of $D$ and $\mathcal{X}_r$, or equivalently, $D_r = D \setminus D_f$, where $D_f = D \bigcap \mathcal{X}_f$. Consequently, any retraining mechanism ($\mathcal{R_T}$) results in a new model with parameters $\theta^r$ whose output distribution $P_{\theta^r}$ seeks to match the distribution $P_{\mathcal{X}_r}$. 

\subsection{Machine Unlearning}
As this retraining is computationally costly, current \emph{machine unlearning} mechanisms ($\mathcal{U_M}$) change the parameters of the model from $\theta^{init}$ to $\theta^u$ so that with very high probability the generative models never give outputs that belong to the domain $\mathcal{X}_f$. Now depending upon the outputs from the retrained model $G_{\theta^r}$ and the unlearned model $G_{\theta^u}$ we define exact weak unlearning \cite{unlearning4} and approximate weak unlearning \cite{unlearning7} as follows

\newtheorem{definition}{Definition}
\begin{definition}
\textbf{(Exact Weak Unlearning)} Given a retrained generative model $G_{\theta^r}(.)$, we say the model $G_{\theta^u}(.)$ is an exact weak unlearned model iff  $\forall z \sim P_{\mathcal{Z}} , \mathcal{O} \subset \mathcal{X}$ the following holds:
\begin{equation}
  \Pr(G_{\theta^r}(z) \in \mathcal{O}) = \Pr(G_{\theta^u}(z) \in \mathcal{O})  
\end{equation}
\end{definition}

\begin{definition}
\textbf{(($\epsilon, \delta$) Approximate Weak Unlearning)} Given a retrained generative model $G_{\theta^r}(.)$, we say the model $G_{\theta^u}(.)$ is an $(\epsilon,\delta)$ approximate weak unlearned model for a given $\epsilon,\delta \geq 0$ iff  $\forall z \sim P_{\mathcal{Z}} , \mathcal{O} \subset \mathcal{X}$ the following conditions hold:
\begin{equation}
\Pr(G_{\theta^r}(z) \in \mathcal{O}) \leq 
e^\epsilon \Pr(G_{\theta^u}(z) \in \mathcal{O})+\delta
\end{equation}
\begin{equation}
    \Pr(G_{\theta^u}(z) \in \mathcal{O}) \leq e^\epsilon \Pr(G_{\theta^r}(z) \in \mathcal{O})+\delta
\end{equation}
\end{definition} 

Definitions 1 and 2 imply that the unlearned model's output should closely resemble that of the naively retrained model. However, it's important to note that retraining or unlearning is a viable option when we have access to the original dataset $D$ and knowledge of the initial model parameters $\theta^{init}$ and architectures. Yet, in many practical scenarios, these generative models are utilized as black boxes for various downstream tasks. In such cases, the conventional unlearning methods become infeasible. When faced with this challenge, where direct unlearning is not an option, we turn to the alternative solution of filtering the outputs generated by these black-box models. The subsequent section will delve into the concept of filtering and elucidate how it effectively translates into a form of weak unlearning for these black-box generative models.

\subsection{Filtering as Weak Unlearning}
\begin{figure}[ht]
    \centering
    \includegraphics[scale=0.4]{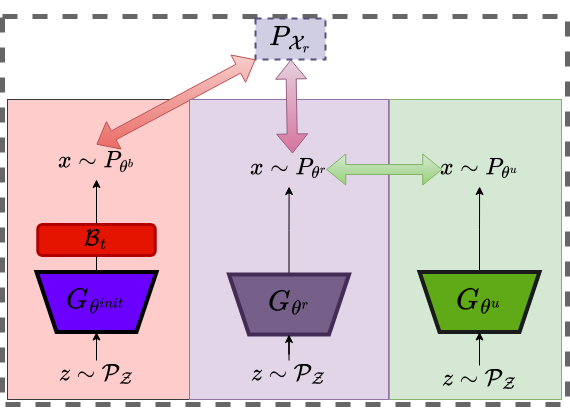}
    \caption{Filtering as Weak Unlearning in Black-Box Generative Models: The left-most block represents the black-box generator with posthoc
    blocking layer with parameters $\theta^b = (\theta^{init}, t)$ whose output distribution $P_{\theta^b}$, the middle block represents a retrained generative model with output distribution $P_{\theta^r}$, the right-most block represents a $(\epsilon,0)$ unlearned model with output distribution $P_{\theta^u}$.}
    \label{fig1}
\end{figure}

In a black-box setting as described, the underlying datasets $D_r$, $D_f$, and the initial model parameter $\theta^{init}$ remain unknown. As illustrated in Fig-\ref{fig1}, our desired filtering model $G_{\theta^b}$ is designed to approximate the distribution $P_{\mathcal{X}_r}$. In a hypothetical scenario where we have access to the model parameters and architecture, any unlearned model $G_{\theta^u}$ would aim to match a completely retrained model's output distribution $P_{\theta^r}$. Leveraging this concept, the following theorem highlights that filtering can be interpreted as a form of weak unlearning for these types of black-box generative models, where the objective is to approximate the distribution in the output space.\\ 

\begin{theorem}
    If there exists a retrained generative model with parameters ${\theta^r}$ such that $d_{ln-TV}( \ln P_{\mathcal{X}_r}|| \ln P_{\theta^r}) \leq \epsilon_1$ and a generative model with posthoc blocking layer denoting parameters ${\theta^b}$ such that $d_{ln-TV}(\ln P_{\mathcal{X}_r}|| \ln P_{\theta^b}) \leq \epsilon_2$ for some $\epsilon_1, \epsilon_2 \geq0$, then for a $(\epsilon,0)-$ approximately weak unlearned model with parameter $\theta^u$ the following holds: $\forall \mathcal{O}\subset\mathcal{X}_r$ 
    \begin{equation}
        |\ln \Pr (G_{\theta^b}(z) \in \mathcal{O}) - \ln \Pr(G_{\theta^u}(z) \in \mathcal{O})| \leq \epsilon + \epsilon_1 + \epsilon_2
    \end{equation}
\end{theorem}

\textbf{\textit{Remarks:}} Here the $d_{ln-TV}(.||.)$ denotes the total variational distance in terms of the log of the distributions. \\

\begin{proof}
    \begin{equation}
  d_{ln-TV}( \ln P_{\mathcal{X}_r}|| \ln P_{\theta^r}) = \sup_{\mathcal{O} \in \mathcal{F}}|\ln P_{\mathcal{X}_r}(\mathcal{O})- \ln P_{\theta^r}(\mathcal{O})| \leq \epsilon_1  
\end{equation}
$$\implies  |\ln P_{\mathcal{X}_r}(\mathcal{O}) - \ln P_{\theta^r}(\mathcal{O})| \leq \epsilon_1 ; \forall \mathcal{O} \in \mathcal{F}$$
$$\implies  {-\epsilon_1} \leq \ln \left(\frac{P_{\mathcal{X}_r}(\mathcal{O})}{P_{\theta^r}(\mathcal{O})}\right) \leq {\epsilon_1} ; \forall \mathcal{O} \in \mathcal{F}$$
As, $\mathcal{F}$ is a $\sigma$-algebra then $\forall x \in \mathcal{X}_f \text{ if the subset } \{x\}\in \mathcal{F}$
$$\implies e^{-\epsilon_1} P_{\theta^r} \leq_{(i)} P_{\mathcal{X}_r} \leq_{(ii)} e^{\epsilon_1} P_{\theta^r} \text{ (Uniformly Bounded)}$$ 
By a similar argument as before,
\begin{equation}
  d_{ln-TV}( \ln P_{\mathcal{X}_r}|| \ln P_{\theta^b}) = \sup_{\mathcal{O} \in \mathcal{F}}|\ln P_{\mathcal{X}_r}(\mathcal{O})- \ln P_{\theta^b}(\mathcal{O})| \leq \epsilon_2 
\end{equation}
$$\implies e^{-\epsilon_2} P_{\theta^b} \leq_{(iii)} P_{\mathcal{X}_r} \leq_{(iv)} e^{\epsilon_2} P_{\theta^b} \text{ (Uniformly Bounded)}$$ 
Now assume an $(\epsilon,0)$-Approximate weak unlearning model $G_{\theta^u}$ then equations (5) and (6) implies $\forall z \sim P_{\mathcal{Z}}$  
\begin{equation}
    |\ln \Pr(G_{\theta^r}(z) \in \mathcal{O}) - \ln \Pr(G_{\theta^u}(z) \in \mathcal{O})| \leq \epsilon
\end{equation}
Now the objective, $$|\ln \Pr (G_{\theta^b}(z) \in \mathcal{O}) - \ln \Pr(G_{\theta^u}(z) \in \mathcal{O})|$$ 
\begin{align*}
     & \leq |\ln \Pr (G_{\theta^b}(z) \in \mathcal{O}) - \ln \Pr (G_{\theta^r}(z) \in \mathcal{O})|_{(A)}  \\ 
     & + |\ln \Pr (G_{\theta^r}(z) \in \mathcal{O}) - \ln \Pr(G_{\theta^u}(z) \in \mathcal{O})|_{(B)}
\end{align*}
By equation (2) and (3) the second term 
\begin{equation}
    (B)= |\ln \Pr (G_{\theta^r}(z) \in \mathcal{O}) - \ln \Pr(G_{\theta^u}(z) \in \mathcal{O})|\leq \epsilon
\end{equation}
Now, using ineq. (ii) and ineq. (iii) we get the below inequalities respectively
\begin{equation}
  \Pr (G_{\theta^r}(z) \in \mathcal{O}) = P_{\theta^r}
    (\mathcal{O}) \geq e^{-\epsilon_1} P_{\mathcal{X}_r}(\mathcal{O})
\end{equation}
\begin{equation}
  \Pr (G_{\theta^b}(z) \in \mathcal{O}) = P_{\theta^b}
    (\mathcal{O}) \leq e^{\epsilon_2} P_{\mathcal{X}_r}(\mathcal{O})
\end{equation}

Taking $\ln$ on both sides of inequality (9) and (10), then subtracting gives
\begin{equation}
  (A)=|\ln \Pr (G_{\theta^b}(z) \in \mathcal{O}) - \ln \Pr (G_{\theta^r}(z) \in \mathcal{O})| \leq \epsilon_1 + \epsilon_2 
\end{equation}
Adding equations (8) and (11) gives
$$|\ln \Pr (G_{\theta^b}(z) \in \mathcal{O}) - \ln \Pr(G_{\theta^u}(z) \in \mathcal{O})| \leq \epsilon + \epsilon_1 + \epsilon_2 $$
\end{proof}

\newtheorem{corollary}{\textbf{Corollary}}[theorem]
\begin{corollary}
    For every $(\epsilon,0)-$ \emph{Approximate Weak Unlearning} algorithm there exists a filtering mechanism such that for some $\epsilon' \geq 0$ the following holds
    \begin{equation}
        \Pr(G_{\theta^b}(z) \in \mathcal{O}) \leq e^{(\epsilon+\epsilon')} \Pr(G_{\theta^u}(z) \in \mathcal{O})
    \end{equation}
    \begin{equation}
        \Pr(G_{\theta^u}(z) \in \mathcal{O}) \leq e^{(\epsilon + \epsilon')} \Pr(G_{\theta^b}(z) \in \mathcal{O})        
    \end{equation}
\end{corollary}

\section{Methodology}
\subsection{Problem Formulation and Method Overview}
As previously mentioned, the pre-trained generator $G_{\theta^{init}}$ is trained using a dataset $D$ comprising $m$ independent and identically distributed (i.i.d) samples denoted as $\{x_i\}_{i=1}^m$, where $x_i \overset{iid}{\sim} P_{\mathcal{X}}$. In our specific scenario, both the dataset $D$ and the initial model parameters $\theta^{init}$ remain undisclosed. Our objective is to formulate a filtering mechanism that relies on user feedback to selectively exclude undesired outputs. In this context, the user is presented with a collection of $r$ samples, denoted as $\mathcal{S} = \{y_i\}_{i=1}^{r}$, where each $y_i$ represents a generated sample originating from the pre-trained GAN. Within this user feedback system, the user's role is to designate a relatively small subset of these samples, denoted as $\mathcal{S}_n = \{y^n_i\}_{i=1}^{s}$, as negative samples. These negative samples correspond to the instances that exhibit undesired features. Additionally, the user selects a portion of the samples, referred to as $\mathcal{S}_p = \{y^p_i\}_{i=1}^{s}$, as positive samples. Positive samples are those that do not possess the undesired features and are considered desirable. It's essential to note that gathering user feedback can be resource-intensive. Therefore, we operate within a low-data regime, where the no of feedback samples marked by the user, denoted as $s$, is very small (constrained to be less than 100) i.e. $s \ll r$. This limitation acknowledges the potential challenges associated with obtaining extensive user feedback and emphasizes the need to devise effective filtering mechanisms even in scenarios with limited data availability.

In this study, we implement a two-stage process to filter out undesired outputs effectively. In the initial stage termed as \emph{Undesired Feature Representation} stage, we encode the representation of the undesired feature within the latent space. This latent space may either be the implicit latent space inherent to the Generative Adversarial Network (GAN) or a latent space that is learnable. Subsequently, in the second step termed as \emph{Similarity Thresholding} stage, we compute the projection similarity between newly sampled latent vectors and this encoded representation of the undesired feature. This similarity calculation serves as the basis for determining the degree of alignment between the latent vectors and the undesired feature. Utilizing this similarity metric, we judiciously apply a thresholding mechanism to separate and exclude outputs that contain the undesired feature.

\subsection{Filtering Mechanism}

\subsubsection{Undesired Feature Representation stage}
In this particular phase of our methodology, the central objective is to uncover a comprehensive and distinct representation of the undesired feature. This representation should exhibit a clear presence in the negative images while being absent in the positive images. This task is inherently challenging due to the intricate and often entangled relationships between various semantic features present within the images. However, as previously studied~\cite{style-gan1,style-gan2,cluster-gan} the latent space, where latent vectors encode critical information about the data, offers a unique and powerful capability for rich and meaningful representation. But it is important to emphasize that the quality and effectiveness of this representation fundamentally depend on the inherent characteristics (e.g. disentanglement, clustering) of the latent space itself. Consequently, the properties, dimensions, and organization of the latent space play a pivotal role in determining the accuracy and utility of the representation we derive.

In the context of our approach, we are provided with two sets of samples: $\mathcal{S}_n = \{y^n_i\}_{i=1}^{s}$ and $\mathcal{S}_p = \{y^p_i\}_{i=1}^{s}$, each containing $s$ samples. From these sets, we extract the corresponding latent vectors, denoted as $\mathcal{Z}_n = \{z^n_i\}_{i=1}^{s}$ and $\mathcal{Z}_p = \{z^p_i\}_{i=1}^{s}$, respectively. This extraction is achieved through a \emph{Latent Projection Function (LPF)}, denoted as $\pi(.)$ such that $\forall i \in \{1,2,\ldots,s\}$; $z_i = \pi(y_i)$. Now, our goal is to represent the unique undesired feature in a concise manner. To do this, we introduce a function called the \emph{Undesired Representation Function (URF)}, denoted as $g(\mathcal{Z}_n,\mathcal{Z}_p)$. This function takes as input the latent vectors $\mathcal{Z}_n$ and $\mathcal{Z}_p$ and produces a distinctive representation of the undesired feature, which we denote as $z_{undesired} = g(\mathcal{Z}_n,\mathcal{Z}_p)$.

\begin{algorithm}[!t]
    \caption{\textbf{FAST:} Feature Aware Similarity Thresholding}\label{alg1}
    \textbf{Required}: positive samples($\{y^p_i\}^s_{i=1}$), negative samples($\{y^n_i\}^s_{i=1}$), test samples($\{y_i\}_{i=1}^t$), LPF($\pi(.)$), URF($g(.)$),
    \begin{algorithmic}
        \STATE{Initialize: $i \gets 1$, $\mathcal{Z}_p \gets \{\}$, $\mathcal{Z}_n \gets \{\}$} 
        \WHILE{$i \leq s$}
            \STATE{$z^p_i \gets \pi(y^p_i)$, $z^n_i \gets \pi(y^n_i)$} 
            \STATE{$\mathcal{Z}_p.\text{append}(z^p_i)$, $\mathcal{Z}_n.\text{append}(z^n_i)$}
        \ENDWHILE
        \STATE{$z_{undesired} \gets g(\mathcal{Z}_p, \mathcal{Z}_n)$}
        \STATE{$T_{th} \gets \frac{1}{2s}\sum_{i=1}^s \left[\text{sim}(z^n_i, z_{undesired}) + \text{sim}(z^p_i, z_{undesired})\right]$}
        \STATE{Initialize: $j \gets 1$, $I_{out} \gets \{\}$}
        \WHILE{$j \leq t$}
            \IF{$\text{sim}(z_j, z_{undesired}) \geq T_{th}$}
                \STATE{$\mathcal{B}_{z_{undesired}}(z_j) \gets 0$}
            \ELSE
                \STATE{$\mathcal{B}_{z_{undesired}}(z_j) \gets 1$, $I_{out}$.append($z_j$)}
            \ENDIF
        \ENDWHILE
        \STATE{\textbf{Output}: $I_{out}$}
    \end{algorithmic}
\end{algorithm}

\subsubsection{Similarity Thresholding stage}
In the context of our work, we draw inspiration from the clustering properties~\cite{cluster-gan} of latent vectors, which encode similar features, tend to be close to each other in the latent space of the generative model. This observation forms the basis for our approach, as we aim to effectively distinguish between latent vectors that correspond to desired and undesired features, ultimately allowing us to filter out the undesired outputs generated by the model. In this step, our approach involves the generation of a set of new latent vectors denoted as $\{z_i\}_{i=1}^t$. These latent vectors are sampled from a latent space distribution, typically following a normal distribution. Each $z_i$ is essentially a numerical representation of a potential data point that the generative model can produce. To assess the alignment between these newly sampled latent vectors and the undesired feature, we introduce a similarity metric, denoted as $\text{sim}(z,z_{undesired})$. This metric quantifies how similar a given latent vector $z$ is to the undesired feature vector $z_{undesired}$ in the latent space. The similarity score is computed as follows.
\begin{equation}
    \text{sim}(z,z_{undesired}) = \frac{z^T \cdot z_{undesired}}{||z_{undesired}||}
\end{equation}
With similarity scores calculated for all latent vectors concerning both positive samples $\{z^p_i\}_{i=1}^{s}$ (those without undesired features) and negative samples $\{z^n_i\}_{i=1}^{s}$ (those containing undesired features), we proceed to establish a similarity threshold denoted as $T_{th}$. This threshold is calculated as below:
\begin{equation}
    T_{th} = \frac{1}{2s} \sum_{i=1}^s \left[\text{sim}(z^n_i, z_{undesired}) + \text{sim}(z^p_i, z_{undesired})\right]
\end{equation}

Now comes the critical decision-making step. To determine which latent vectors should be filtered out due to the presence of undesired features, we use the filtering decision function $\mathcal{B}_{T_{th}}(z_i)$ as follows.
\begin{equation}
    \mathcal{B}_{T_{th}}(z_i) =
  \begin{cases}
    0 & \text{when } \text{sim}(z_i,z_{undesired}) \geq T_{th} \\
    1 & \text{when } \text{sim}(z_i,z_{undesired}) < T_{th}  \\
\end{cases}
\end{equation}
If the similarity between a latent vector $z_i$ and the undesired feature vector $z_{undesired}$ is greater than or equal to the calculated threshold $T_{th}$ indicates that this latent vector should be filtered out, as it contains a significant component of the undesired feature. Conversely, if the similarity falls below the threshold, it signifies that this latent vector can be retained, as it does not contain the undesired feature. Now, it becomes evident that this decision is contingent upon the value of $z_{undesired}$, and this value, in turn, is subject to the specific choices made for the \emph{Latent Projection Function (LPF)} and the \emph{Undesired Representation Function (URF)}.

\begin{figure*}[!t]
    \centering
    \includegraphics[width=\textwidth]{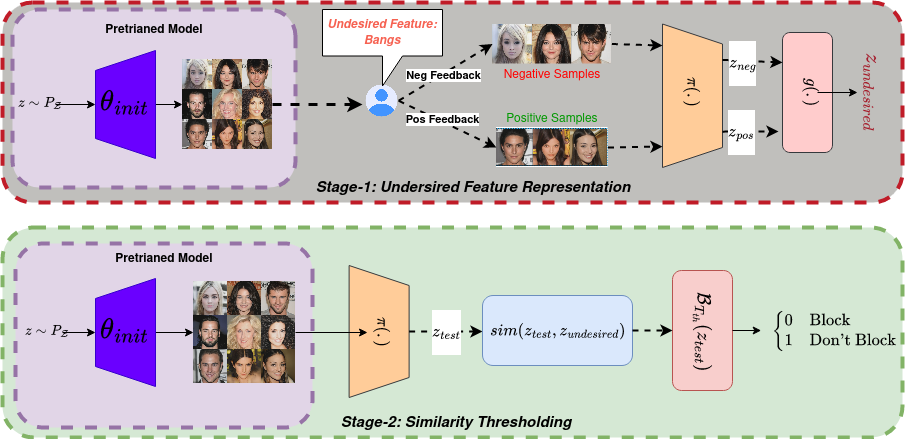}
    \caption{\emph{FAST} filtering mechanism: In stage-1, the undesired latent feature is identified using only a few positive and negative samples marked by the user. The positive and negative samples are projected into the latent space using the \emph{Latent Projection Function} ($\pi(.)$), and subsequently, the undesired feature is retrieved via the \emph{Undesired Representation Function} ($g(.)$). In stage-2, during the inference phase, new test samples are projected into the latent space, and the similarity of their projection with the undesired feature obtained from stage-1 is measured to filter out the negative samples.}
    \label{fig:filtering-mech}
\end{figure*}

\subsection{Choices of LPF and URF}
A crucial aspect to consider in our methodology is the flexibility and adaptability of the \emph{Latent Projection Function (LPF)} and the \emph{Undesired Representation Function (URF)}. \emph{LPF} plays a vital role in mapping the data from its original space into the latent space, where different semantic features can be effectively encoded. On the other hand, \emph{URF} is responsible for crafting a representation of the undesired feature within the latent space. These functions hold a pivotal role in the overall effectiveness of our approach, and selecting them appropriately is essential to ensure the method's success. It's worth emphasizing that there isn't a one-size-fits-all solution; rather, the choice of these functions should be made based on the specific characteristics and demands of the problem under consideration, making \emph{FAST} a unified family of methods.

Here we have analyzed two different LPFs that map into two different latent spaces as follows:

\begin{itemize}
    \item \emph{Implicit Latent Space (Imp-LS):} This is the original latent space, which is the domain of the original GAN ($G_{\theta^{init}}$). Here $\pi(.) = G_{\theta^{init}}^{-1}$.
    \item \emph{Inverted Latent Space (Inv-LS):} This is the inverted latent space ($\mathcal{Z}_{inv}$) obtained through learning an inverter ($I_{\theta^{init}}$) via reconstruction loss such that $I_{\theta^{init}}: \mathcal{X} \rightarrow \mathcal{Z}_{inv}$. Here $\pi(.) = I_{\theta^{init}}$.
\end{itemize}

Further, we take two different URFs that represent the undesired features as follows:

\begin{itemize}
    \item \emph{Mean Difference (MD):} This represents the undesired features as the difference between the means of latent vectors corresponding to negative and positive examples:
    \begin{equation}
        g(\mathcal{Z}_n, \mathcal{Z}_p) = \frac{1}{|\mathcal{Z}_n|} \sum_{i \in \mathcal{Z}_n} z^n_i - \frac{1}{|\mathcal{Z}_p|} \sum_{i \in \mathcal{Z}_p} z^p_i
    \end{equation}

    \item \emph{SVM Normal (Norm-SVM):} This represents the undesired features as the normal vector of a hyperplane obtained after training an SVM with latent vectors from $\mathcal{Z}_n$ and $\mathcal{Z}_p$:
    \begin{equation}
        g(\mathcal{Z}_n, \mathcal{Z}_p) = \textit{SVM-Normal}(\mathcal{Z}_n, \mathcal{Z}_p)
    \end{equation}
\end{itemize}

\subsection{Latent Augmentation} The effectiveness of the proposed filtering mechanism is dependent on the limited feedback samples provided by the user. Due to the scarcity of user feedback samples, this method is prone to overfitting to the few available samples. To mitigate the overfitting issue and enhance the effectiveness of our method, we incorporate latent sampling in the latent space. Formally, we augment the number of negative and positive features in the latent space by sampling 5000 features from their respective empirical distributions, as given below:
\begin{align*}
z^p_{aug} &\sim \mathcal{N}(\Bar{z}^p, \Bar{\Sigma}p) & \hspace{4mm}
z^n_{aug} &\sim \mathcal{N}(\Bar{z}^n, \Bar{\Sigma}_n)
\end{align*}
Here, $\Bar{z}^p$, $\Bar{\Sigma}_p$, $\Bar{z}^n$, $\Bar{\Sigma}_n$ represent the mean and covariance matrix of the positive and negative latent vectors, respectively. 



\section{Experiments and Results}
\subsection{Datasets and Models}
As previously mentioned, the primary objective of filtering is to prevent the display of samples that exhibit specific undesired features. In this context, we explore two distinct unlearning settings:
\begin{itemize}
    \item \emph{Class-level filtering:} For this setting, we utilize the MNIST dataset (LeCun et al., 1998), which comprises 60,000 black and white images of handwritten digits having dimensions of 28 × 28. Aiming to achieve class-level filtering,  we have used pre-trained DC-GAN on MNIST. Specifically, we focus on filtering out two digit classes: 5 and 8.
    \item \emph{Feature-level filtering:} In this scenario, we turn to the CelebA-HQ dataset (Liu et al., 2015), which contains 30,000 high-quality RGB celebrity face images with dimensions of 256×256. For this, we have taken state-of-the-art Style-GAN2 pre-trained on CelebA-HQ. Here, we target the feature-level filtering of subtle features, specifically (a) bangs and (b) hats.
\end{itemize}

\subsection{Training Details:}

\subsubsection{Pre-trained GAN} 
As specified, we utilize two pre-trained GAN architectures: DC-GAN trained on MNIST and Style-GAN2 trained on CelebA-HQ. The DC-GAN architecture and weights are obtained from an open-source repository \footnote{\href{https://github.com/csinva/gan-vae-pretrained-pytorch}{Source Code: Pre-trained DC-GAN trained on MNIST dataset}}, while Style-GAN2 architecture is sourced from another repository \footnote{\href{https://github.com/rosinality/Style-GAN2-pytorch}{Source Code: Pre-trained Style-GAN2 trained on CelebA-HQ dataset}}. To obtain the pre-trained GAN on CelebA-HQ, images are resized to $256\times256$ to conform to the Style-GAN2 architecture. The latent space dimensions are set to $100\times1$ for DC-GAN and $512\times1$ for Style-GAN2. GAN training employs non-saturating adversarial loss with path-regularization. Default optimizers and hyperparameters provided in the code are utilized for training. The training duration for CelebA-HQ is $3.6\times10^5$ epochs.

\subsubsection{Pre-trained Classifier}
\label{sec:app-clf-details} We leverage pre-trained classifiers to emulate user feedback, obtaining positive and negative samples by evaluating generated samples from the pre-trained GAN through these classifiers. Additionally, these classifiers simulate human filtering, and in the evaluation metrics section, we compute FID, density, and coverage based on this classifier's output.
We use simple LeNet model\footnote{\href{https://github.com/csinva/gan-vae-pretrained-pytorch/tree/master/mnist\_classifier}{Source Code: Pre-trained Classifier trained on MNIST dataset}} for classification among different digits of MNIST dataset. The model is trained with a batch-size of $256$ using Adam optimizer with a learning rate of $2\times10^{-3}$, $\beta_1 = 0.9$ and  $\beta_2 = 0.999$. The model is trained for a resolution of $32\times32$ same as the pre-trained GAN for $12$ epochs. After training the classifier has an accuracy of $99.07\%$ on the test split of the MNIST dataset.
Further for feature level-filtering, we use ResNext50 model~\cite{resnext50} for classification among different facial attributes contained in CelebA-HQ\footnote{\href{https://github.com/rgkannan676/Recognition-and-Classification-of-Facial-Attributes/tree/main}{Source Code: Pre-trained Classifier trained on CelebA-HQ dataset}}. 
The classifier is trained with a batch-size of $64$ using Adamax optimizer with a learning rate of $2\times10^{-3}$, $\beta_1 = 0.9$ and  $\beta_2 = 0.999$. The model is trained for a resolution of $256\times256$ for $10$ epochs. We also employ image augmentation techniques such as horizontal flip, image resize, and cropping to improve the performance of the classifier. The trained model exhibits a test accuracy of $91.93\%$.

\begin{table*}[!t]
\caption{Recall($\uparrow$), AUC($\uparrow$), FID ($\downarrow$) Density($\uparrow$) and Coverage($\uparrow$) after filtering \textbf{MNIST} classes with only \textbf{20 Pos. and 20 Neg.} user-feedback}
\centering
\resizebox{\textwidth}{!}{
\begin{tabular}{l|ccccc|ccccc}
\toprule
\multicolumn{1}{c|}{\multirow{2}{*}{\textbf{Methods}}} & \multicolumn{5}{c|}{\textbf{Class-5}} & \multicolumn{5}{c}{\textbf{Class-8}} \\
 & \multicolumn{1}{c}{\textbf{Recall}} & \multicolumn{1}{c}{\textbf{AUC}} & \multicolumn{1}{c}{\textbf{FID}} & \multicolumn{1}{c}{\textbf{Density}} & \multicolumn{1}{c|}{\textbf{Coverrage}} & \multicolumn{1}{c}{\textbf{Recall}} & \multicolumn{1}{c}{\textbf{AUC}} & \multicolumn{1}{c}{\textbf{FID}} & \multicolumn{1}{c}{\textbf{Density}} & \multicolumn{1}{c}{\textbf{Coverage}} \\
\midrule
\textbf{Base Classifier}  & 0.19$\pm$0.01 & 0.45$\pm$0.01 & 3.60$\pm$0.14 & 0.93$\pm$0.01 & 0.91$\pm$0.01 & 0.35$\pm$0.02 & 0.53$\pm$0.01 & 1.42$\pm$0.07 & \textbf{0.99$\pm$0.01} & 0.99$\pm$0.00 \\
\textbf{Base Classifier+Data Aug.}  & 0.10$\pm$0.00 & 0.50$\pm$0.00 & 0.80$\pm$0.03 & 0.93$\pm$0.00 & \textbf{0.99$\pm$0.00} & 0.10$\pm$0.00 & 0.50$\pm$0.00 & 0.55$\pm$0.07 & \textbf{0.99$\pm$0.00} & \textbf{1.01$\pm$0.00} \\
\textbf{Imp-LS+MD}  & \textbf{0.69$\pm$0.06} & 0.69$\pm$0.04 & 2.34$\pm$0.97 & 0.95$\pm$0.02 & 0.96$\pm$0.02  & \textbf{0.71$\pm$0.03} & \textbf{0.72$\pm$0.03} & 1.78$\pm$0.33 & 0.98$\pm$0.01 & 0.97$\pm$0.01 \\
\textbf{Imp-LS+MD+Latent Aug.}  & \textbf{0.69$\pm$0.06} & 0.69$\pm$0.04 & 2.30$\pm$0.94 & \textbf{0.96$\pm$0.02} & 0.96$\pm$0.01  & \textbf{0.71$\pm$0.03} & \textbf{0.73$\pm$0.03} & 1.79$\pm$0.36 & \textbf{0.99$\pm$0.01} & 0.97$\pm$0.01 \\
\textbf{Imp-LS+Norm-SVM}  & 0.67$\pm$0.11 & 0.69$\pm$0.03 & 2.27$\pm$1.00 & \textbf{0.96$\pm$0.01} & 0.96$\pm$0.02 & 0.69$\pm$0.04 & 0.71$\pm$0.02 & 1.90$\pm$0.36 & \textbf{0.99$\pm$0.02} & 0.97$\pm$0.01 \\
\textbf{Imp-LS+Norm-SVM+Latent Aug.}  & 0.56$\pm$0.02 & 0.57$\pm$0.05 & 3.05$\pm$0.72 & 0.92$\pm$0.01 & 0.95$\pm$0.01 & 0.60$\pm$0.10 & 0.63$\pm$0.06 & 2.39$\pm$0.37 & \textbf{0.99$\pm$0.01} & 0.96$\pm$0.01 \\
\textbf{Inv-LS+MD} & 0.11$\pm$0.02 & 0.74$\pm$0.04 & 0.65$\pm$0.95 & 0.93$\pm$0.02 & \textbf{0.99$\pm$0.02} & 0.01$\pm$0.01 & 0.61$\pm$0.08 & 0.64$\pm$0.08 & \textbf{0.99$\pm$0.01} & \textbf{1.10$\pm$0.00} \\
\textbf{Inv-LS+MD+Latent Aug.} & 0.12$\pm$0.02 & 0.75$\pm$0.03 & \textbf{0.63$\pm$0.97} & 0.93$\pm$0.02 & \textbf{0.99$\pm$0.01} & 0.01$\pm$0.00 & 0.59$\pm$0.06 & 0.58$\pm$0.08 & \textbf{0.99$\pm$0.01} & 0.99$\pm$0.01 \\
\textbf{Inv-LS+Norm-SVM}  & 0.55$\pm$0.02 & \textbf{0.83$\pm$0.04} & 0.72$\pm$1.01 & \textbf{0.96$\pm$0.01} & 0.97$\pm$0.02  & 0.11$\pm$0.14 & 0.68$\pm$0.04 & 0.85$\pm$0.33 & \textbf{0.99$\pm$0.00} & 0.99$\pm$0.01 \\
\textbf{Inv-LS+Norm-SVM+Latent Aug.}  & 0.30$\pm$0.02 & 0.70$\pm$0.04 & 2.22$\pm$1.02 & 0.93$\pm$0.01 & 0.94$\pm$0.02  & 0.12$\pm$0.14 & \textbf{0.72$\pm$0.05} & \textbf{0.48$\pm$0.03} & \textbf{0.99$\pm$0.01} & 0.99$\pm$0.01 \\
\bottomrule
\end{tabular}
\label{tab:mnist-filtering-results}
}
\end{table*}

\begin{table*}[ht]
\caption{ Recall($\uparrow$), AUC($\uparrow$), FID ($\downarrow$) Density($\uparrow$) and Coverage($\uparrow$) after filtering \textbf{Celeba-HQ} features with only \textbf{20 Pos. and 20 Neg.} user-feedback}
\centering
\resizebox{\textwidth}{!}{
\begin{tabular}{l|ccccc|ccccc}
\toprule
\multicolumn{1}{c|}{\multirow{2}{*}{\textbf{Methods}}} & \multicolumn{5}{c|}{\textbf{Class-Bangs}} & \multicolumn{5}{c}{\textbf{Class-Hats}} \\
 & \textbf{Recall} & \textbf{AUC} & \textbf{FID} & \textbf{Density} & \textbf{Coverage} & \textbf{Recall} & \textbf{AUC} & \textbf{FID} & \textbf{Density} & \textbf{Coverage} \\
\midrule
\textbf{Base Classifier}  & 0.20$\pm$0.01 & 0.69$\pm$0.01 & 10.58$\pm$0.42 & \textbf{1$\pm$0.01} & 0.88$\pm$0.01  & 0.27$\pm$0.01 & 0.45$\pm$0.01 & 6.37$\pm$0.51 & \textbf{1.08$\pm$0.02} & 0.99$\pm$0.00 \\
\textbf{Base Classifier+Data Aug.} & 0.10$\pm$0.00 & 0.50$\pm$0.00 & \textbf{0.08$\pm$0.02} & 0.99$\pm$0.01 & \textbf{1$\pm$0.00}  & 0.1$\pm$0.00 & 0.5$\pm$0.00 & 6.57$\pm$0.04 & 1.05$\pm$0.01 & \textbf{1$\pm$0.00} \\
\textbf{Imp-LS+MD}  & 0.91$\pm$0.05 & 0.90$\pm$0.04 & 3.71$\pm$1.12 & 0.99$\pm$0.02 & 0.94$\pm$0.02 & 0.77$\pm$0.06	& 0.84$\pm$0.02 & 3.55$\pm$1.01 & 0.98$\pm$0.01 & 0.93$\pm$0.01 \\
\textbf{Imp-LS+MD+Latent Aug.}  & 0.91$\pm$0.07 & 0.89$\pm$0.04 & 3.69$\pm$0.97 & 0.99$\pm$0.02 & 0.94$\pm$0.02 & 0.78$\pm$0.06	& 0.85$\pm$0.03	& 3.50$\pm$0.92&	0.98$\pm$0.01 &	0.93$\pm$0.01 \\
\textbf{Imp-LS+Norm-SVM}  & \textbf{0.93$\pm$0.06} & 0.92$\pm$0.02 & 2.54$\pm$1.00 & 0.99$\pm$0.01 & 0.96$\pm$0.02 & 0.82$\pm$0.04	& 0.86$\pm$0.02	& 3.81$\pm$0..93	& 0.95$\pm$0.01	& 0.93$\pm$0.01 \\
\textbf{Imp-LS+Norm-SVM+Latent Aug.} & \textbf{0.93$\pm$0.02} & \textbf{0.93$\pm$0.05} & 2.31$\pm$0.72 & 0.99$\pm$0.01 & 0.97$\pm$0.01 &0.77$\pm$0.03	&0.84$\pm$0.05	&3.35$\pm$0.65	&0.99$\pm$0.15	&0.94$\pm$0.01 \\
\textbf{Inv-LS+MD} & 0.89$\pm$0.01 & 0.89$\pm$0.01 & 9.91$\pm$1.14 & 0.96$\pm$0.01 & 0.85$\pm$0.01 & \textbf{0.86$\pm$0.03	}& 0.88$\pm$0.02 & 9.06$\pm$0.72 & 0.94$\pm$0.01 & 0.83$\pm$0.02 \\
\textbf{Inv-LS+MD+Latent Aug.} & 0.89$\pm$0.01 & 0.89$\pm$0.01 & 9.86$\pm$1.04 & 0.96$\pm$0.01 & 0.85$\pm$0.01 & \textbf{0.86$\pm$0.03} & \textbf{0.89$\pm$0.02} &	9.05$\pm$0.70 &	0.94$\pm$0.01 &	0.83$\pm$0.02 \\
\textbf{Inv-LS+Norm-SVM}  & \textbf{0.94$\pm$0.02} & \textbf{0.93$\pm$0.03} & 3.61$\pm$0.65 & \textbf{1.01$\pm$0.02} & 0.96$\pm$0.01 &	0.74$\pm$0.01 &	0.85$\pm$0.01 &	5.15$\pm$0.81 &	0.93$\pm$0.02	& 0.90$\pm$0.01 \\
\textbf{Inv-LS+Norm-SVM+Latent Aug.} & 0.92$\pm$0.02 & \textbf{0.93$\pm$0.02} & 3.83$\pm$0.70 & 1.00$\pm$0.01 & 0.94$\pm$0.01 &	0.78$\pm$0.01 &	0.86$\pm$0.01 &	5.48$\pm$0.75 &	0.94$\pm$0.02 &	0.90$\pm$0.01 \\
\bottomrule
\end{tabular}
\label{tab:celebahq-filtering-results}
}
\end{table*}

\subsection{Baseline and Evaluation Metrics}
\subsubsection{Baseline} To the best of our knowledge, there exists no established baseline method for effectively filtering out undesired output samples generated by a black-box GAN. So to evaluate the effectiveness of our method, we implement a fundamental yet essential baseline strategy of using a binary classifier designed to leverage user feedback consisting of positive and negative samples. This classifier termed as \emph{Base Classifier} is tailored to identify and filter out undesired output samples, serving as a foundational baseline in our approach. To implement this \emph{Base Classifier} we have used ResNext50 model~\cite{resnext50} pre-trained on Imagenet as the base feature extractor on top of which 3-layer MLP trained with binary cross entropy loss, fine-tuned for both MNIST and CelebA-HQ datasets. Adamax optimizer is used for training, running for 200 epochs with a learning rate of 0.07. Given the limited user feedback ($\leq 100$ samples), we introduce data augmentation techniques including rotation, blur, perspective, and auto-augment to boost the performance of this base classifier. Each image is augmented into 64 variations during training. 

\subsubsection{Evaluation Metrics}
\label{evaluation}
To gauge the effectiveness of our proposed techniques and the baseline methods, we utilize three fundamental evaluation metrics:
\begin{enumerate}
    \item \textbf{Recall:} In this task, recall is a crucial metric as it measures the effectiveness of filtering out negative images generated by the model. It quantifies the proportion of negative images successfully filtered. An optimal recall value of 1 indicates that all negative images have been correctly filtered out.
    \item \textbf{AUC:} The AUC score, derived from varying the threshold $T_{th}$ and calculating the area under the curve of the True Positive rate ($P_D$) vs. the False Positive rate ($P_F$), quantitatively assesses the filtering method's effectiveness. Thus AUC captures the effect of varying threshold. A maximum AUC score of 1 indicates optimal performance.
    \item \textbf{Fr\'echet Inception Distance (FID), Density and Coverage:} While the earlier mentioned metrics quantify the effectiveness of filtering, they do not assess the quality of the filtered output distribution in comparison to a distribution obtained through human-level filtering. To replicate human-level filtering, we utilize a pre-trained classifier trained on labeled training data\footnote{Please note that the original dataset is unavailable during the \emph{FAST} filtering process. Consequently, the use of the original dataset is solely for evaluation purposes with the \emph{Pre-trained Classifier}.}. To address this aspect, we calculate the Fréchet Inception Distance (FID)~\cite{fid} between the generated samples after the FAST method and a pre-trained classifier trained on labeled data. A lower FID value signifies a closer match to the pre-trained classifier's output distribution. Additionally, we employ Density and Coverage~\cite{density-coverage} metrics to evaluate how the output of our proposed filtering method deviates from the distribution obtained with a pre-trained (human-level) classifier. The optimal value for Coverage is 1, and higher values of Density and Coverage signify better filtering performance.
\end{enumerate}

\subsection{Results}
Table-\ref{tab:mnist-filtering-results} and Table-\ref{tab:celebahq-filtering-results} provide evaluations of different filtering methods on the MNIST and CelebA-HQ datasets, respectively. We observe that our proposed methods give better Recall and AUC values for both datasets while comparable to superior performance for other metrics.

The analysis focuses on two specific class features, class-5 and class-8 for MNIST, and subtle features of bangs and hats for CelebA-HQ, each with only 20 positive and 20 negative user feedback samples. For both MNIST and CelebA-HQ, we observe that the \emph{Base Classifier}, trained on the user feedback data, exhibits moderate performance in both classes. However, incorporating data augmentation significantly improves accuracy, with notable enhancements in the AUC score. Methods employing \emph{Implicit-Latent Space (Imp-LS)} with \emph{ Mean Difference (MD)} or \emph{Normal SVM (Norm-SVM)} feature representation-based filtering mechanism (\emph{Imp-LS+MD/Norm-SVM}) demonstrate better performance than the \emph{Base Classifier} in terms of Recall and AUC scores for both the datasets. The \emph{Imp-LS+MD} and \emph{Imp-LS+MD+Latent-Augmentation} methods demonstrate superior recall performance, surpassing the \emph{Base Classifier} by approximately 3.5 times for MNIST class-5 and 2 times for MNIST class-8. Additionally, these methods achieve remarkable recall scores, outperforming the \emph{Base Classifier} by almost 4.1 times for CelebA-HQ bangs and 1.7 times for hats features, respectively. The combination of \emph{Inversion-Latent Space (Inv-LS)} with \emph{Normal SVM (Inv-LS+Norm-SVM)}  feature representation-based mechanism yields superior AUC scores, demonstrating improvements of 66.4\% and 35.8\% over the \emph{Base Classifier} for MNIST class-5 and class-8, respectively. In terms of FID, Density, and Coverage we observe that our proposed family of methods gives superior to comparative performance with respect to the \emph{Base Classifier}.

These findings suggest the efficacy of the \emph{FAST} family of methods. Although no single method achieves superior results across all metrics, in most cases, \emph{Inv-LS+Norm-SVM} and \emph{Inv-LS+Norm-SVM+Latent Aug.} perform significantly better than other methods for both Recall and AUC scores. This makes them the most desirable methods among the proposed ones.

\subsection{Ablation Study}
\subsubsection{Effect of user-feedback size}
Table-\ref{lab:length.vs.scores} presents an evaluation of recall and AUC scores, examining the efficacy of diverse methods on two distinct features—MNIST class-5 and CelebA-HQ bangs—across varying numbers of positive and negative feedback samples. 

Notably, the \emph{Imp-LS+MD} and \emph{Imp-LS+MD+Latent Augmentation} filtering methods consistently exhibit competitive recall and AUC scores, underscoring their robust performance across both datasets. The influence of altering the number of feedback samples becomes apparent in the results, with certain methods showcasing improved performance as the number of positive and negative samples increases. For instance, the \emph{Inv-LS+Norm-SVM} method achieves a recall of 0.48 and 0.72 for MNIST class-5 with 40 and 60 samples, respectively. This observation highlights the sensitivity of the method to the feedback sample size, emphasizing the significance of considering the quantity of feedback samples when evaluating filtering methods.

\begin{table}[!t]
\caption{Evaluation of Recall and AUC scores with varying no of pos and negative feedback}
\centering
\resizebox{\columnwidth}{!}{
\begin{tabular}{cl|cc|ll}
\toprule
\multirow{2}{*}{\textbf{No. of (Pos. + Neg.) Samples}} & \multicolumn{1}{c|}{\multirow{2}{*}{\textbf{Methods}}} & \multicolumn{2}{c|}{\textbf{MNIST Class-5}} & \multicolumn{2}{c}{\textbf{CelebA-HQ - Bangs}} \\ 
 & & \textbf{Recall} & \textbf{AUC Scores} & \textbf{Recall} & \textbf{AUC Scores} \\
\midrule
\multirow{8}{*}{(40 + 40)} & \textbf{Imp-LS+MD} & 0.68$\pm$0.07 & 0.72$\pm$0.02 & 0.90$\pm$0.01 & 0.92$\pm$0.04 \\
 & \textbf{Imp-LS+MD+Latent Aug.} & 0.68$\pm$0.07 & 0.72$\pm$0.03 & 0.89$\pm$0.07 & 0.91$\pm$0.04 \\
 & \textbf{Imp-LS+Norm-SVM} & 0.65$\pm$0.06 & 0.71$\pm$0.04 & 0.95$\pm$0.05 & 0.95$\pm$0.02 \\
 & \textbf{Imp-LS+Norm-SVM+Latent Aug.} & 0.56$\pm$0.05 & 0.59$\pm$0.03 & 0.94$\pm$0.02 & 0.94$\pm$0.05 \\
& \textbf{Inv-LS+MD} & 0.33$\pm$0.02 & 0.75$\pm$0.04 & 0.93$\pm$0.02 & 0.91$\pm$0.02 \\
& \textbf{Inv-LS+MD+Latent Aug.} & 0.32$\pm$0.02 & 0.74$\pm$0.03 & 0.94$\pm$0.01 & 0.91$\pm$0.01 \\
& \textbf{Inv-LS+Norm-SVM} & 0.49$\pm$0.02 & 0.77$\pm$0.04 & 0.93$\pm$0.01 & 0.95$\pm$0.01 \\
& \textbf{Inv-LS+Norm-SVM+Latent Aug.} & 0.20$\pm$0.02 & 0.66$\pm$0.04 & 0.94$\pm$0.02 & 0.95$\pm$0.02 \\
\midrule
\multirow{8}{*}{(60 + 60)} & \textbf{Imp-LS+MD} & 0.69$\pm$0.02 & 0.75$\pm$0.01 & 0.90$\pm$0.01 & 0.93$\pm$0.04 \\
 & \textbf{Imp-LS+MD+Latent Aug.} & 0.68$\pm$0.02 & 0.75$\pm$0.01 & 0.91$\pm$0.06 & 0.92$\pm$0.03 \\
 & \textbf{Imp-LS+Norm-SVM} & 0.66$\pm$0.04 & 0.73$\pm$0.00 & 0.92$\pm$0.04 & 0.94$\pm$0.02 \\
 & \textbf{Imp-LS+Norm-SVM+Latent Aug.} & 0.54$\pm$0.04 & 0.61$\pm$0.02 & 0.95$\pm$0.02 & 0.95$\pm$0.05 \\
& \textbf{Inv-LS+MD} & 0.39$\pm$0.02 & 0.76$\pm$0.04 & 0.90$\pm$0.02 & 0.91$\pm$0.04 \\
& \textbf{Inv-LS+MD+Latent Aug.} & 0.40$\pm$0.04 & 0.76$\pm$0.03 & 0.90$\pm$0.04 & 0.91$\pm$0.03 \\
& \textbf{Inv-LS+Norm-SVM} & 0.72$\pm$0.02 & 0.78$\pm$0.01 & 0.96$\pm$0.01 & 0.97$\pm$0.01 \\
& \textbf{Inv-LS+Norm-SVM+Latent Aug.} & 0.32$\pm$0.02 & 0.81$\pm$0.01 & 0.97$\pm$0.02 & 0.97$\pm$0.02 \\
\bottomrule
\end{tabular}
\label{lab:length.vs.scores}
}
\end{table}


\subsubsection{Evaluation under no positive feedback}
\begin{figure}[!ht]
    \centering
    \subfloat[pos. mining Class-5]{\label{fig:pos. mining Class-5}\includegraphics[width=0.2\textwidth]{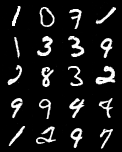}}~
    \subfloat[pos. mining Class-8]{\label{fig:pos. mining Class-8}\includegraphics[width=0.2\textwidth]{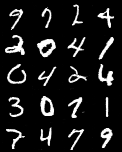}}~
\caption{Results of \emph{positive mining} given only 20 negative samples by the user for MNIST dataset. Samples are generated by the pre-trained GAN after obtaining the positive samples through \emph{positive mining}.}
\label{fig:pos_mine}
\end{figure}

In our endeavor to enhance the adaptability of our approach, we extend its functionality to accommodate scenarios where users provide exclusively negative samples without any positive instances. In such cases, we employ a technique referred to as \emph{positive mining} to infer positive samples. Specifically, given the user identified negative samples $\mathcal{S}_n = \{y^n_i\}_{i=1}^{s}$, their corresponding negative latents $\mathcal{Z}_n = \{z^n_i\}_{i=1}^{s}$ are computed. The negative similarity scores set $SS(z^n_i,\Bar{z}^n) = \{sim(z^n_i,\Bar{z}^n) : z^n_i \in \mathcal{S}_n, \Bar{z}^n = \frac{1}{s}\sum^s_{i=1} z^n_i \}$ is defined. Now given a newly sampled latent $z \sim P_\mathcal{Z}$, it is a positive latent if $sim(z,\Bar{z}^n) \ll \min_i SS(z^n_i,\Bar{z}^n)$. The results of positive mining are given in Fig.~\ref{fig:pos_mine} for the MNIST dataset, using only 20 negative samples from Class-5 and Class-8. These results were generated by a pre-trained GAN after obtaining positive noises through the \emph{positive mining} approach. It is evident that positive mining not only selects noise samples from other classes but also maintains diversity across these classes, demonstrating its effectiveness even when user feedback is limited to negative samples.

Table-\ref{lab:only-neg-evaluation} offers a comprehensive evaluation summary under the constraint of 20 negative and 0 positive feedback instances for MNIST class-5 and CelebA-HQ bangs datasets. Leveraging positive mining, our methods, \emph{Imp-LS+MD} and \emph{Imp-LS+MD+Latent Augmentation}, achieve noteworthy recall values of 0.72 and 0.73 for MNIST class-5, and 0.98 and 0.99 for CelebA-HQ bangs, respectively. Remarkably, these recall values surpass the performance achieved when both positive and negative feedback samples are available.

\begin{table}[ht]
\caption{Evaluation of Recall and AUC scores with only 20 negative and 0 positive feedback}
\centering
\resizebox{\columnwidth}{!}{
\begin{tabular}{l|cc|cc}
\toprule
\multicolumn{1}{c|}{\multirow{2}{*}{\textbf{Methods}}} & \multicolumn{2}{c|}{\textbf{MNIST Class-5}} & \multicolumn{2}{c}{\textbf{CelebA-HQ - Bangs}} \\
 & \textbf{Recall} & \textbf{AUC Scores} & \textbf{Recall} & \textbf{AUC Scores} \\
\midrule
\textbf{Imp-LS+MD} & 0.72$\pm$0.09 & 0.70$\pm$0.01 & 0.98$\pm$0.005 & 0.88$\pm$0.04 \\
\textbf{Imp-LS+MD+Latent Aug.} & 0.73$\pm$0.11 & 0.72$\pm$0.03 & 0.99$\pm$0.007 & 0.89$\pm$0.04 \\
\textbf{Imp-LS+Norm-SVM} & 0.66$\pm$0.07 & 0.71$\pm$0.04 & 0.97$\pm$0.006 & 0.91$\pm$0.02 \\
\textbf{Imp-LS+Norm-SVM+Latent Aug.} & 0.64$\pm$0.11 & 0.64$\pm$0.05 & 0.98$\pm$0.002 & 0.90$\pm$0.05 \\
\textbf{Inv-LS+MD} & 0.26$\pm$0.02 & 0.67$\pm$0.02 & 0.96$\pm$0.004 & 0.89$\pm$0.02\\
\textbf{Inv-LS+MD+Latent Aug.} & 0.25$\pm$0.02 & 0.66$\pm$0.05 & 0.96$\pm$0.005 & 0.89$\pm$0.03 \\
\textbf{Inv-LS+Norm-SVM} & 0.39$\pm$0.04 & 0.70$\pm$0.04 & 0.97$\pm$0.003 & 0.94$\pm$0.02 \\
\textbf{Inv-LS+Norm-SVM+Latent Aug.} & 0.28$\pm$0.02 & 0.69$\pm$0.03 & 0.98$\pm$0.002 & 0.93$\pm$0.05 \\
\bottomrule
\end{tabular}
\label{lab:only-neg-evaluation}
}
\end{table}

Furthermore, \emph{Imp-LS+Norm-SVM} and \emph{Imp-LS+Norm-SVM+Latent Augmentation} demonstrate competitive performance in terms of recall and AUC scores across both datasets, showcasing the robustness of our approach. Conversely, the \emph{Inversion-Latent Space (Inv-LS)} methods exhibit lower recall values, underscoring the differential impact of various methods under the specified feedback conditions. This ability to operate effectively with only negative feedback demonstrates the versatility and practicality of our approach in real-world scenarios with diverse user interactions.

\section{Conclusion, Limitations and Future Work}
To deploy deep generative models safely and responsibly, it is crucial to eliminate outputs with undesired features. In this research, we explored algorithms to prevent generating such samples from a black-box pre-trained Generative Adversarial Network (GAN). Recognizing that current unlearning methods are inadequate due to the lack of access to the pre-trained model's parameters, we propose a filtering methodology as an alternative solution. In our theoretical framework, we establish that filtering can be viewed as a form of weak unlearning. Thus to effectively filter undesired output, we propose an algorithm named \emph{Feature Aware Similarity Thresholding (FAST)}, which suppresses unwanted outputs by systematically encoding the representation of undesired features in the latent space. Notably, our methodology works effectively even when the original training dataset and model parameters are hidden from the user, making it applicable to zero-shot settings. This work is a pioneering effort that highlights the intricate relationship between filtering and unlearning, offering a new perspective on the challenges posed by black-box generative models.

\par \textbf{Limitations:} The primary inspiration for our method is the understanding that latent space efficiently encodes semantic features. Therefore, our goal is to represent the undesired feature within the latent space, which necessitates certain useful properties such as disentanglement between different features. Disentanglement ensures that distinct features are represented independently, allowing for precise filtering of undesired attributes. However, in many cases, the latent space may be entangled due to the intricate interplay of semantic features within images. This entanglement can lead to the representation of multiple correlated features in a combined manner, making it challenging to isolate and filter out only the undesired features. As a result, our methodology might inadvertently filter out correlated features along with the undesired features in such scenarios.
\par \textbf{Future works:} Acknowledging the limitations, concerns, and further applicability of our method, we have identified key aspects to address in future work: (i) \emph{Learning disentangled latent space:} We plan to explore the use of inversion mechanisms to learn a more disentangled latent space, which would reduce the effect of feature entanglement. This approach aims to improve the precision of filtering undesired features without inadvertently affecting correlated features. (ii) \emph{Application of FAST to black-box diffusion models:} We aim to extend the \emph{FAST} algorithm to black-box diffusion models. We hypothesize that further modifications to the \emph{FAST} mechanism will be required for diffusion models because these models generate samples through multiple Langevian backward steps. This process suggests that the initial latent space may not encode semantic features effectively, necessitating additional investigation and adaptation of the \emph{FAST} algorithm to ensure its applicability and effectiveness in this context.

\bibliography{main}
\bibliographystyle{IEEEtran.bst}
\end{document}